\pdfoutput=1

\documentclass[11pt]{article}

\usepackage[]{acl}

\usepackage{times}
\usepackage{latexsym}

\usepackage[T1]{fontenc}

\usepackage[utf8]{inputenc}

\usepackage{microtype}

\usepackage{booktabs}
\usepackage{amsmath}
\usepackage{amssymb}
\usepackage[switch]{lineno} 
\usepackage{multirow}
\usepackage{floatrow}
\usepackage{boldline}
\usepackage{makecell}
\usepackage{graphicx}
\newfloatcommand{capbtabbox}{table}[][\FBwidth]
\usepackage[labelfont=bf]{caption}

\title{Then and Now: Quantifying the Longitudinal Validity of Self-Disclosed Depression Diagnoses}

\author{Keith Harrigian \and Mark Dredze\\
  Johns Hopkins University \\
  \texttt{kharrigian@jhu.edu}, \texttt{mdredze@cs.jhu.edu}}

\begin{document}
\maketitle


\begin{abstract}
Self-disclosed mental health diagnoses, which serve as ground truth annotations of mental health status in the absence of clinical measures, underpin the conclusions behind most computational studies of mental health language from the last decade. However, psychiatric conditions are dynamic; a prior depression diagnosis may no longer be indicative of an individual's mental health, either due to treatment or other mitigating factors. We ask: to what extent are self-disclosures of mental health diagnoses actually relevant over time? We analyze recent activity from individuals who disclosed a depression diagnosis on social media over five years ago and, in turn, acquire a new understanding of how presentations of mental health status on social media manifest longitudinally. We also provide expanded evidence for the presence of personality-related biases in datasets curated using self-disclosed diagnoses. Our findings motivate three practical recommendations for improving mental health datasets curated using self-disclosed diagnoses:
\begin{enumerate}
    \item Annotate diagnosis dates and psychiatric comorbidities
    \item Sample control groups using propensity score matching
    \item Identify and remove spurious correlations introduced by selection bias
\end{enumerate}
\end{abstract}


\section{Introduction} \label{sec:Introduction}


The ability to provide equitable access to psychiatric healthcare has become more difficult than ever, inhibited by an entanglement of lingering public policy effects \cite{miranda2020policy}, heightened levels of physician burnout \cite{johnson2018mental}, and infrastructural challenges arising from global crisis \cite{davis2021distance}. Meanwhile, social media platforms have become the predominant means of communication for much of the population, providing the opportunity to share personal experiences and seek support from others \cite{mueller2021demographic}. Noting these parallel timelines, computational scientists have devoted substantial effort to engineering statistical models capable of translating social media data into reliable insights regarding mental health. Core objectives of this work include optimizing psychiatric treatment, identifying early stages of mental illness, and measuring the effect of public policy on a population's well-being \cite{losada2017erisk,fine2020assessing}.

The most significant advances in computational mental health research have not come from improved modeling architectures \cite{benton2017multitask}, but from methods for curating large-scale datasets which contain robust and clinically-relevant ground truth annotations of mental health status \cite{coppersmith2014quantifying}. Use of regular expressions to identify genuine self-disclosures of a psychiatric diagnosis remains one of the most widely adopted annotation mechanisms by the research community \cite{chancellor2020methods,harrigian2021state}, offering a relatively reliable proxy in place of clinical measures which are not only costly to collect, but also often unable to be shared beyond a single institution due to patient privacy policies \cite{macavaney2021community}. Datasets leveraging self-disclosed diagnoses as annotations of mental health status have yielded a variety of insights that align with clinical knowledge and psychological theory \cite{mowery2017understanding,lee2021micromodels}. However, a growing body of work has raised questions about whether such datasets provide sufficient information to train statistical models that generalize to new populations \cite{harrigian2020models,aguirre2021gender}.


Despite the prevalence of datasets dependent on self-disclosure, no analyses have considered how associating a single self-disclosed diagnosis label with data from a variable-length period of time may inhibit the learning of robust statistical relationships. If a user tweets a depression diagnosis in 2015, is their data from 2018 still representative of the condition? Presentation of several mental health conditions change dynamically and (sometimes) precipitously over time \cite{collishaw2004time}. Yet, it remains common in the computational research community to treat mental health conditions as a static attribute with equal relevance at multiple time points \cite{macavaney2018rsdd}. In reality, it is likely that only a small fraction of an individual's social media activity is appropriate for training optimal classifiers. Moreover, that a mental health status label may be appropriate for only a subset of time suggests that evaluations of longitudinal model generalization as they are traditionally structured in the community may be insufficient \cite{sadeque2018measuring}.

We ask: to what extent do mental health diagnosis self-disclosures remain valid over time? We focus specifically on extended durations (i.e., multiple years), a setting which has particular relevance to those who wish to estimate generalization strength of their statistical classifiers for use in longitudinal monitoring applications, as well as those interested in updating existing models with new data to mitigate the effects covariate shift \cite{agarwal2021temporal}. In reviewing recent online activity from individuals in the 2015 CLPsych Shared Task dataset who disclosed a depression diagnosis on Twitter over five years ago \cite{coppersmith2015clpsych}, we not only acquire a new understanding of how presentations of mental health status on social media present over time, but also find new evidence to support prior claims regarding the presence of personality-related confounds in datasets curated using self-disclosures \cite{preoctiuc2015role,vukojevic2021pandora}. Our analysis provides critical guidance to practitioners as they curate mental health social media datasets, while also elucidating factors which inhibit robustness in a dataset that remains one of the most widely adopted by the research community.


\section{Background} \label{sec:Background}

The majority of mental health research based on social media leverages the same experimental design---assume individuals have a fixed mental health status and attempt to infer this latent attribute using historical online activity traces (e.g., posts, follower network dynamics) \cite{guntuku2017detecting,chancellor2020methods}. This training setting is convenient given the inherent complexities of acquiring temporally-granular psychiatric measures at scale \cite{canzian2015trajectories}. However, the setting implicitly relies on assumptions that are not supported by clinical knowledge regarding psychiatric dynamics \cite{johnson2002dynamical,schoevers2005depression}. Some work has been done to incorporate time-based priors into mental health models, which allow practitioners to train statistical classifiers using a static label while also explicitly accounting for longitudinal variation in label relevance \cite{wongkoblap2019predicting,uban2021understanding}. Others have eschewed the use of a static label altogether and instead curated datasets that contain multiple points of ground truth mental health status, albeit still with some element of historical data aggregation \cite{chancellor2016quantifying}.

Temporally-aware classifiers have achieved better performance benchmarks than their static counterparts in some cases \cite{rao2020mgl}, though these evaluations remain limited by the dearth of data with mental health status annotations at multiple time points. Meanwhile, datasets which do support dynamic evaluation are curated almost exclusively using protected clinical measures \cite{reece2017forecasting}, cost-intensive interviews \cite{nobles2018identification}, or non-trivial shifts in non-language-based online behavior \cite{de2016discovering}. 

Computational studies that have focused on self-disclosed diagnoses have not comprehensively reviewed how individual activity evolves over long periods of time \cite{saha2021understanding}. Our study thus fulfills an important void in the research space by providing a new understanding of long term mental health dynamics in social media, and more particularly, within convenience samples curated using self-disclosed diagnoses.



\section{Data} \label{sec:Data}

\begin{table}[t]
    \centering
    \begin{tabular}{c c c c}
        \textbf{Dataset} & \textbf{Dates}  & \textbf{\# Users} &  \textbf{\# Posts} \\ \toprule
        Original & 
            2012 -- 2015 & 
            \begin{tabular}{@{}l@{}}D: 477\\C: 872\end{tabular} &
            \begin{tabular}{@{}l@{}}D: 1,121,388\\C: 1,907,508\end{tabular} \\ \midrule
        Updated & 
            2012 -- 2021 & 
            \begin{tabular}{@{}l@{}}D: 444\\C: 172\end{tabular} &
            \begin{tabular}{@{}l@{}}D: 1,372,868\\C: 546,826\end{tabular} \\
    \end{tabular}
    \caption{Summary statistics for the original and updated versions of the 2015 CLPsych Shared Task dataset, further stratified by [C]ontrol and [D]epression groups.}
    \label{tab:datasets}
\end{table}

We support our study using a newly updated version of the 2015 CLPsych Shared Task dataset \cite{coppersmith2015clpsych}. The original Twitter dataset was constructed in a two-stage process, with regular-expressions first being used to identify candidate self-disclosures of a depression diagnosis and experts manually verifying the authenticity of the match thereafter. Individuals in the control group were sampled randomly from the 1\% public Twitter stream such that the joint distribution of inferred age and gender attributes \cite{sap2014developing} was in alignment with the depression group. Up to 3,000 tweets were acquired for each individual in the resulting sample using Twitter's public API. The dataset has not only become one of the most widely adopted social media datasets for mental health \cite{harrigian2021state}, but also inspired the annotation procedures for numerous successors across various platforms and languages \cite{cohan2018smhd,shen2018cross}.

In line with guidance from \citet{benton2017ethical}, individual identifiers in the official version of the CLPsych dataset have been anonymized, with linkages between anonymized and de-anonymized identifiers erased in entirety. However, the original de-anonymized identifiers remain available under explicit permission from \citet{coppersmith2015clpsych}, who provided this information to reverse engineer the original anonymization mapping. To do so, we first query up to 3,200 of the most recent tweets from each de-anonymized user identifier using Twitter's public API and further isolate all relevant tweets found in our institution's cache of Twitter's 1\% data stream. We identify candidate pairs of anonymized and de-anonymized accounts based on overlap of raw timestamps within the original dataset's collection window. Normalized text (i.e., punctuation removal, case standardization) from candidate pairs is compared using exact matching to verify final linkages.

Statistics for the original dataset and its updated counterpart are provided in Table \ref{tab:datasets}. We find that a majority of accounts which were unable to be linked had significantly smaller activity traces in the original dataset. These accounts are likely to either have been deleted in entirety or to have tweeted with a small enough frequency such that the 1\% stream does not contain any samples. The discrepancy in match rates between individuals in the depression and control groups is unfortunately not fully-understood, though discussions with the dataset's authors suggest this may just be an artifact of the original archival process.

\textbf{Preprocessing.} Twitter's language tags and automatic language identification \cite{lui2012langid} are used to isolate English text. Retweets are excluded to most acutely highlight personal experiences with depression over time. Unless specified otherwise, keyword-based tweet filtering is applied to preemptively mitigate sampling-induced biases which can artificially inflate estimates of predictive performance. Some of these biases have been recognized and addressed by the research community (e.g., filtering tweets which include diagnosis disclosures and/or mental health related keywords/hashtags) \cite{de2014mental}, while others have been traditionally overlooked.

A preliminary qualitative analysis of influential $n$-grams and their source tweets reveals a previously unrecognized surplus of ``fan accounts'' (e.g., supporters of Harry Styles and Demi Lovato) and tweets containing account statistics (e.g., new followers) within the depression cohort. Meanwhile, daily horoscope tweets were identified with an anomalous frequency within the control group. The latter two sources of noise do not have a clear clinical explanation, while the former (i.e., fan accounts) arises in the context of discussion regarding the mental health of young celebrities. Although some of these motifs represent genuine behavioral correlates of depression, their importance in prediction tends to be inflated due to context of the original collection time period.


\section{Inference Under Latent Dynamics} \label{sec:InferenceDynamics}

Enabling reliable use of statistical models to evaluate change in mental health status remains a core objective for computational researchers \cite{choi2020development,fine2020assessing}. Our success in this task domain critically depends on access to ground truth at multiple time points, not only for evaluating generalization error \cite{demasi2017meaningless,tsakalidis2018can}, but also for mitigating the effects of covariate shift \cite{sugiyama2012machine}. As discussed above, it is often trivial to update activity traces for individuals with a prior mental health diagnosis disclosure. Nonetheless, clinical knowledge suggests original disclosure-based labels may not be relevant over the course of time, either due to a condition's episodic presentations \cite{angst2009long} or the effects of psychiatric treatment \cite{saha2021understanding}. We ask whether the CLPsych Shared Task dataset supports this theory.


\textbf{Methods.} A natural framework for answering this inquiry emerges from computational research regarding label noise \cite{frenay2013classification}. Under such a perspective, we can view changes in mental health status as a stochastic process which blindly alters the correctness of class labels over time. The implications of this mechanism allow us to reason about predictive performance of a statistical classifier within and outside of the time period in which it is trained. Differences in within-time-period performance for two different time periods may be caused by two factors---different levels of label noise and/or different signal-to-noise ratios. Meanwhile, degradation in performance when transferring a classifier from one time period to another may be caused by three possible factors---label noise in the source time period, label noise in the target time period, or distributional shift between the time periods. Although isolated differences in predictive performance in a longitudinal setting do not implicate a single causal factor, multiple comparisons taken together may allow us to reason about underlying changes in the data.

This logic guides our search for evidence in support of the hypothesis that mental health annotations cannot be treated as fixed attributes. We consider a standard longitudinal domain transfer setup \cite{huang2019neural}, chunking the CLPsych dataset into three discrete three-year periods\footnote{Time periods were chosen to maximize the number of discrete windows while ensuring enough posts were available to construct informative individual-level representations.} (2012--2015, 2015--2018, 2018--2021) and evaluating within- and between-time-period predictive performance for all available pairs. We use Monte Carlo Cross Validation \cite{xu2001monte} to obtain estimates of predictive generalization, chosen over alternative protocols that would be unreliable given the limited sample size of the updated CLPsych dataset \cite{varoquaux2018cross}.

Each iteration of the cross validation procedure (1,000 total) begins by randomly splitting individuals into a 60/40 train/test split, with control and depression groups demographically aligned\footnote{Aligned on gender and age dimensions.} using propensity scores \cite{imbens2015causal}. To control for differences in data availability between time periods, we not only constrain the sampling process such that splits have an \emph{equal class balance}, but also that individual-level representations are constructed using an \emph{equal document history size} (250 randomly-sampled posts from each time period). A single binary logistic regression classifier provided with document-term TF-IDF representations \cite{baeza1999modern} is fit for each time period using data from individuals in the training set. Each classifier is applied to all three time periods, evaluating performance using individuals in the sampled test set.

\begin{table}[t]
    \centering
    \begin{tabular}{c c c c}
        & \multicolumn{3}{c}{\textbf{Test}} \\
        \textbf{Train} & 2012-2015 & 2015-2018 & 2018-2021 \\ \toprule
            2012-2015 & 
                $.71_{(.70, .72)}$ & 
                $.66_{(.65, .66)}$ & 
                $.69_{(.68, .70)}$ \\
            2015-2018 & 
                $.66_{(.65, .67)}$ &
                $.66_{(.65, .66)}$ &
                $.68_{(.67, .69)}$ \\
            2018-2021 &
                $.65_{(.65, .66)}$ &
                $.67_{(.66, .68)}$ &
                $.68_{(.67, .69)}$
    \end{tabular}
    \caption{Average test-set area under the curve (AUC) and 95\% confidence intervals across 1,000 Monte Carlo Cross Validation iterations. Within-time-period performance is significantly higher around the original disclosure window than in subsequent time periods.}
    \label{tab:performance}
\end{table}

\textbf{Results.} We report the average test set area under the curve (AUC) and 95\% confidence intervals for each discrete time period pairing in Table \ref{tab:performance}. Focusing first on \emph{within-time-period} performance (top left to bottom right diagonal), we find that within-time-period performance is significantly higher in the dataset's original time period (2012-2015) than within subsequent time periods. This holds true even when running experiments only with individuals that have sufficiently-sized post histories in the new time periods, demonstrating that the outcome is not an artifact of survivor bias. At a high level, the differences in within-time-period performance suggest that either label noise has increased or that the signal-to-noise ratio has decreased over time.

%


Unfortunately, examination of \emph{between-time-period} generalization does not conclusively resolve which of these two factors are responsible for the variation. Focusing first on models trained using data from older time periods (top right triangle), we do not observe any significant difference in predictive performance compared to the benchmarks established by models trained and deployed during the same time period. This serves as a contrast to models deployed on older data (bottom left triangle), where we note that classifiers trained on both of the new time periods incur a loss when being applied to the original CLPsych dataset time period. Interestingly, the absolute differences in performance are minimal. We note that the coefficients of the logistic regression classifiers from each independent time period exhibit significantly positive Pearson correlations, ranging from 0.47 to 0.52, and in turn promote stable performance.


\textbf{Discussion.} Although these experiments have not conclusively answered our primary research question regarding longitudinal label validity, they have provided evidence that not all time periods of data are equally informative for training a robust depression classifier. Critically, these results suggest that practitioners cannot assume it better to train a depression classifier using new data, which may be more relevant to their deployment scenario, if it means potentially compromising the temporal relevance of the original ground truth annotations.

What remains to be understood is \emph{why} the predictive task appears to become more difficult in the updated time periods at a statistically significant level, but not one that would necessarily raise immediate concerns to a practitioner. Had underlying dynamics significantly changed since the original data collection period, we would have expected to see a more dramatic loss in predictive performance. Has the mental health status for these individuals genuinely remained static, or is there a spurious confound in the data inflating our performance estimates?



\section{Interpreting Model Performance} \label{sec:Qualitative}

We attempt to better understand the variation in predictive performance estimated above by comparing language within the updated dataset to the original CLPsych sample. In particular, we adopt a mixed methods approach that allows us to estimate changes in the proportion of depression labels which remain relevant in the updated dataset, and to qualitatively summarize drivers of model decision-making across time periods. We support our analysis by manually coding content-related motifs within a large sample of document histories in the updated dataset, focusing primarily on criteria for diagnosing depression as defined within the DSM-5 \cite{american2013diagnostic}. We draw inspiration from the growing literature on ``train-set debugging'' \cite{koh2017understanding,han2020explaining}, which leverages instance attribution and other diagnostics to succinctly interpret the relationship between training data, learned model parameters, and downstream predictions.


\textbf{Methods.} An annotator is presented with up to 30 anonymized tweets made by a single individual during one of the time periods and asked to indicate whether the individual exhibits evidence of depression. The annotator must mark one of four options --- Uncertain, No Evidence, Some Evidence (Moderate Confidence), Strong Evidence (High Confidence). Explicit disclosures of a depression diagnosis and references to living with depression are automatically assigned to the Strong Evidence category. Otherwise, the annotator is instructed to indicate their confidence based on the nine DSM-5 criteria for diagnosing depression \cite{american2013diagnostic} and their prior knowledge regarding the presentation of mental health conditions within social media. If at least some evidence of a depression diagnosis is indicated, the annotator is asked to identify whether the depression appears to be in remission (e.g., discussion of overcoming depression). They are also asked to indicate which DSM-5 criteria and/or prior knowledge was used to inform their decision, along with any other notable thematic content.


Our goal of this analysis is \emph{not} to make diagnostic claims regarding the mental health status of individuals in our dataset, but rather to broadly understand what the statistical classifiers are learning. Accordingly, tweets presented to the annotator are those which had the largest positive effect on the classifier's estimated probability of depression, as measured by their influence on user-level predictions within a given time period $\tau$. Formally, we define the influence of a tweet $I(x)$ amongst a set of tweets $x \in X_{\tau}$ as follows:
\begin{align*}
    I(x) = \sum_{k=1}^{K} P_{k,\tau}(y=1|X_\tau) - P_{k,\tau}(y=1|X_{\tau}^{\neg{x}})
\end{align*}
where $P_{k,\tau}(\cdot)$ is the probability of depression estimated by a classifier trained on the $k$-th random sample of data from time period $\tau$, out of $K$ total samples. As was the case in the classification experiments above, each training sample contains 60\% of the available data, with the learned classifiers only being applied to the remaining 40\% of individuals at each iteration. We refrain from filtering mental health related tweets and those containing explicit diagnosis disclosures, as the goal in this experiment is not to quantify predictive ability, but rather to identify evidence of depression over time. Note that we control for distributional shift over time by estimating influence using a model trained during the time period in which a tweet was posted.

\textbf{Data.} A total of 300 individuals (574 total instances) were selected randomly for annotation. One author, a doctoral student in computer science with multiple years experience working with the CLPsych dataset, was responsible for all coding. They consulted one additional co-author, an expert in computational modeling of social media and mental health, to develop a common mental model for identifying DSM-5 criteria and other common linguistic motifs in the text. During a pilot round of coding, 16 thematic patterns were identified within the annotated instances to complement the original DSM-5 criteria. Exemplary tweets (paraphrased non-trivially to preserve anonymity \cite{ayers2018don}) for each of the DSM-5 criteria and alternative thematic categories are provided in Appendix \ref{apx:interpretingevidence}. A breakdown of annotation results is presented in the Table \ref{tab:annotations}.  We provide a distribution of the top 20 most common evidence categories amongst individuals who displayed at least some evidence of a depression diagnosis in Appendix \ref{apx:interpretingevidence}.

\textbf{Reliability.} Two non-authors with a background in computational psychology independently annotated a subsample of the coded instances to assess the primary coder's reliability. Agreement regarding whether an individual exhibits evidence of depression was fair to moderate; we observe Krippendorff's $\alpha$ measures of $0.438$ and $0.499$  for the four-class (Uncertain, No Evidence, Some Evidence, Strong Evidence) and three-class (Uncertain, No Evidence, Some or Strong Evidence) scenarios, respectively \citep{krippendorff2011computing}. Agreement regarding remission status varied significantly between pairs of annotators and was generally weaker than agreement regarding evidence of depression ($\alpha=0.356$). We include an analysis of the disagreements in Appendix \ref{apx:intepretreliability} to better contextualize observations from the primary coder's annotations. Succinctly, we identify two reasons for the variation: 1) each annotator's propensity to select the ``Uncertain'' category, and 2) each annotator's sensitivity to displays of emotion as an indicator of depression.


\begin{table}[t]
    \centering
    \begin{tabular}{c c c c c c}
     &
        \begin{tabular}{@{}c@{}}\textbf{Dates}\end{tabular} &
        \begin{tabular}{@{}c@{}}\textbf{Total}\end{tabular} &
        \begin{tabular}{@{}c@{}}\textbf{Some}\\\textbf{Evi.}\end{tabular} &
        \begin{tabular}{@{}c@{}}\textbf{Strong}\\\textbf{Evi.}\end{tabular} &
        \begin{tabular}{@{}c@{}}\textbf{Not}\\\textbf{Active}\end{tabular} \\ \toprule
        \multirow{3}{*}{\rotatebox[origin=c]{90}{\textbf{Con.}}} & 
        2012-2015 & 
        83 &
        15 & 
        3 & 
        1 \\
    &
        2015-2018 & 
        50 &
        10 & 
        2 & 
        0 \\
    & 
        2018-2021 & 
        40 &
        5 & 
        0 & 
        0 \\ \midrule
        \multirow{3}{*}{\rotatebox[origin=c]{90}{\textbf{Dep.}}} & 
        2012-2015 & 
        215 &
        164 & 
        136 & 
        10 \\
    &
        2015-2018 & 
        107 &
        49 & 
        28 & 
        2 \\
    & 
        2018-2021 & 
        79 &
        31 & 
        16 & 
        1 \\
    \end{tabular}
    \caption{Breakdown of coding labels as a function of time period and labels from the original CLPsych dataset. Clinically aligned evidence of a depression diagnosis becomes less prevalent over time.}
    \label{tab:annotations}
\end{table}

\subsection{What proportion of labels in the updated sample remain relevant?}

In line with underlying clinical knowledge regarding the dynamic nature of depression, we observe a significant decrease in linguistic evidence of depression over the course of time. Roughly 76\% of individuals in the original depression group displayed at least some clear evidence of a depression diagnosis during the first time period (2012-2015), in comparison to 45\% and 39\% of individuals in the 2015-2018 and 2018-2021 time periods, respectively. Across all time periods, only a small number of affirmative instances of depression appear to be in remission. That said, the non-zero level of inactive depression annotations in the original time period highlights an important consideration for practitioners who would like to leverage disclosure-based mechanisms to annotate mental health data moving forward.

The presence of evidence for a depression diagnosis in a subset of the original control group is quite striking. Other studies have raised questions regarding the possible risk of introducing such label noise when curating a control group using a random sampling protocol \cite{wolohan2018detecting}, though none have provided tangible evidence of this contamination to the best of our knowledge. We see that approximately 4\% of individuals in the control group display strong evidence of a depression diagnosis within the original time period. Although relatively small, it is an important reminder of the pitfalls of random control group sampling for health-related social media modeling tasks.

\textbf{Discussion.} The decrease in evidence of a depression diagnosis over time lends support to the introduction of label noise in the updated dataset. Furthermore, it would explain the decrease in predictive performance observed in our previous classification experiments. However, the proportional drop in evidence of a depression diagnosis over time appears too large given the relatively minor reduction in classification accuracy. 


We identify two possible explanations for this inconsistency. First, we recognize the possibility that our annotation procedure is insufficient to provide an annotator with appropriate information and comprehensive criteria for indicating evidence of a depression diagnosis. Only a small subset of an individual's entire post history is displayed to the annotator, a subset chosen using an inherently error-prone statistical ranking method. It is possible that stronger indicators of a depression diagnosis lie outside the 30-tweet sample size window for some individuals. Moreover, the annotator was instructed to rely predominantly on DSM-5 criteria to inform their decision, though several prior computational studies have shown language informative of depression may stray from explicit diagnostic criteria and be difficult for humans to recognize altogether (e.g., increased personal pronoun usage \cite{holtzman2017meta}).


More concerning is the possible presence of non-trivial confounds introduced by the original dataset's sampling/annotation procedure which may artificially inflate predictive performance estimates. Similar types of bias have been identified in prior work when attempting to transfer statistical mental health models trained using proxy-based annotations to new populations of individuals (e.g., demographics, patient populations) \cite{ernala2019methodological,aguirre2021gender}. Although sampling-based artifacts may be causally-related to the original diagnosis disclosure (e.g., a coping mechanism that becomes a hobby, heightened levels of neuroticism), they may be serve as a red herring in place of primary indicators of depression.

\subsection{Do presentations of depression provide evidence of sampling-related confounds?}

Personality-related attributes are prominent features in all periods of the updated dataset. For example, indications of a depressed and/or irritable mood were the most common form of evidence in support of an individual having a depression diagnosis. In many cases, anger and irritation were displayed in the form of interpersonal confrontation (passively and actively) with other Twitter profiles. Negative emotions such as loneliness, fear, and existential dread were also displayed readily amongst those showing signs of a depression diagnosis. This result aligns with knowledge regarding the relationship between personality and depression, with elevated levels of neuroticism (negative affectivity and vulnerability to stress) being common in those living with depression \cite{bagby2008personality,lahey2009public,bondy2021neuroticism}. Although etiologically relevant, this heightened level of emotional affect emerges as one possible artifact which may confound displays of depression and serve as a nuisance variable in linguistic models of the condition \citep{tackman2019depression}.

We also found it common for individuals to mention comorbid psychiatric conditions---such as obsessive compulsive disorder, bipolar disorder, and general anxiety. Many of these conditions share similar underlying symptoms and causes with depressive disorders \cite{franklin2001posttraumatic,goodwin2015overlap}, but tend to assume a different temporal profile \cite{schoevers2005depression}. The significant overlap often makes it difficult for trained physicians to properly diagnose individuals \cite{bowden2001strategies} and for language-based algorithms to achieve appropriate discriminative sensitivity \cite{ive-etal-2018-hierarchical}. We recognize the possibility that these comorbid conditions are active during the updated time periods for some individuals and may assume a proxy role in place of depression.


Although not captured by any single evidence category in isolation, there emerged a distinct propensity for ``oversharing'' amongst individuals from the original dataset's depression group. More specifically, we identified ample discussion of topics that are typically considered socially inappropriate in public discourse spaces (e.g., sexual activity, familial conflict, use of controlled substances). On one hand, this is an interesting finding given that individuals living depression often demonstrate lower levels of emotional self-disclosure \citep{wei2005adult,kahn2009emotional}. On the other hand, we note that prior work in clinical psychology has recognized a similar propensity for depressed and anxious individuals to engage in oversharing within social media \citep{radovic2017depressed,law2020might}.

The theory behind the latter is that social media offers an opportunity to  discuss the oft stigmatized challenges of mental health \cite{betton2015role} and increase feelings of connectedness in a less personal environment \citep{luo2020self}. With this in mind, perhaps it is not surprising that those who have openly disclosed their experience with depression also feel comfortable discussing the aforementioned ``taboo'' topics. Nonetheless, this personal comfort remains relatively unique amongst the larger social media population. The unfortunate effect of this nuance is that it transforms the primary depression inference task into, essentially, a topic-classification task.

\textbf{Discussion.} Our analysis affirms what other recent studies on proxy-based mental health annotations have claimed --- individuals who disclose a mental health condition systematically differ from the larger population of individuals living with that condition \cite{ernala2019methodological,saha2021understanding}. As a research community, we must be careful to disambiguate 1) training a language classifier to identify individuals who live with a mental health condition, and 2) training a language classifier to identify individuals who live with a mental health condition \emph{and} disclose their diagnosis. Inappropriately equating the two creates an opportunity to erroneously estimate population-level dynamics \cite{amir2019mental} and ignore underrepresented voices from communities who tend to possess conservative ideologies regarding mental health \cite{loveys2018cross,aguirre2021gender}.

\section{Discussion} \label{sec:Discussion}

Demand for computational methods to quantify mental health dynamics within social media data is at an all time high \cite{galea2020mental}. However, the potential impact of these methods remains bounded by the robustness of datasets used for their development. Spanning nearly a decade of online activity, our study uniquely identifies evidence of these limitations as they currently manifest in non-clinically derived mental health social media datasets. This evidence leads us to offer three recommendations for enhancing data curation and model evaluation.

\textbf{Annotate Diagnosis Date \& Comorbidities.} We identified several instances within our dataset where a diagnosis disclosure was made in reference to a condition that had since entered remission. In other cases, depression diagnoses were either supplanted by or augmented with alternative psychiatric diagnoses. Indicators regarding the time a diagnosis was made, many of which can be identified using inexpensive algorithms \cite{macavaney2018rsdd}, can provide important signal regarding the temporal relevance of a psychiatric diagnosis. Meanwhile, inclusion of comorbidities may provide researchers an opportunity to model psychiatric heterogeneity \cite{arseniev2018type} and interpret longitudinal generalization.

\textbf{Sample Control Groups using Propensity Matching}. Control group selection is influential in both training and evaluation of statistical models of mental health \cite{pirina2018identifying}. Prior work has leveraged a myriad of criteria to match individuals who have disclosed a psychiatric diagnosis with suitable counterparts---demographics \cite{coppersmith2014quantifying}, online behavior \cite{cohan2018smhd}, and language \cite{de2016discovering}. Though use of inconsistent matching criteria is less than ideal, the absence of any protocol is potentially more problematic \cite{shen2018cross,wolohan2018detecting}. We recommend practitioners leverage propensity-based matching \cite{imbens2015causal} to reduce the effect of self-disclosure biases (e.g., personality, interests, demographics). In addition to the aforementioned dimensions, researchers may augment their criteria using classifiers to infer relevant latent attributes \cite{preoctiuc2015role} or neural models to derive user-level embeddings \cite{amir2017quantifying}.

\textbf{Identify and Filter Sampling Biases.} Our analysis benefited from context that emerged when attempting to train classifiers that generalize over long time periods. However, access to supplementary data is not necessary to understand whether artifacts may exist in a dataset. Algorithmic approaches, such as those from \citet{le2020adversarial}, may be used to identify instances containing spurious correlations. These approaches should be used to augment insights derived from manual annotation and review. We found our technique for ranking the influence of individual posts on user-level predictions began yielding insights after only a few dozen examples, though alternative ranking methodologies are available \cite{uban2021understanding}. Outcomes should be used to inform preprocessing decisions, construct fair evaluations \cite{poliak2018hypothesis}, and inform the description of a dataset within documentation/datasheets \cite{gebru2021datasheets}.

\subsection{Limitations and Qualifiers}

Though our analysis identified data attributes that may inhibit statistical generalization, we also found evidence in support of the validity of self-disclosed diagnoses for annotating mental health status. The majority of individuals within the CLPsych dataset's original time window showed clear evidence of depression that aligns with clinical criteria. Many of these indicators remained stable over the course of time. Moreoever, the 2015 CLPsych Shared Task dataset is just one of many resources in this research community, all of which are likely to exhibit varying degrees of noise depending on their respective sampling protocols. Conclusive statements regarding the validity of self-disclosed diagnoses require evidence from multiple social media platforms, cultural groups, and time periods. 




\section{Ethical Considerations} \label{sec:Ethics}

Ethical challenges emerging from use of public social media data to analyze an individual's mental health have been examined extensively by members of both computational and clinical/public health communities \cite{conway2016social,chancellor2019taxonomy}. Privacy-related concerns are the most poignant for our study, which relies both on de-anonymizing records from a vulnerable population and manually reviewing/analyzing individual posts.

Indeed, many individuals who publicly discuss their mental health or disclose a psychiatric condition within social media admit that they worry about harmful repercussions of sharing such sensitive information with the public \cite{ford2019public,naslund2019risks}. Primary fears include risking occupational stability, damaging interpersonal relationships, and being subjected to hostile communications. Whether potential positive outcomes (e.g., development of systems for recommending mental health care, fiduciary aid to address population-level crises) offset these threats remains largely dependent on an individual's personal life experience. For example, psychiatric patients have expressed stronger approval toward analysis of their social media than members of the general public \cite{mikal2017investigating}. The same holds true amongst younger individuals \cite{naslund2019risks}.

Recognizing these viewpoints, we are careful to mitigate privacy-related risks to the greatest extent possible given our primary research aim. For example, account identifiers distributed within the 2015 CLPsych Shared Task dataset are de-anonymized only temporarily to link updated records with existing post histories. We also redact account handles and URLs from the text analyzed during our manual coding procedure (\S \ref{sec:Qualitative}). In line with protocols enumerated by \citet{benton2017ethical}, all data is stored on a remote server and secured using OS-level group permissions. We perform our analysis under the external guidance of clinical psychologists and psychiatrists. Our study is also reviewed by our Institutional Review Board (IRB), obtaining exempt status under 45 CFR \S 46.104.

Critically, our intention is not to develop a public-facing system for algorithmic analysis of mental health. Rather, our goal is to evaluate the validity of an existing and widely-adopted data curation practice \cite{chancellor2020methods,harrigian2021state}. Failure to comprehensively understand biases that arise under this methodology can have severe detrimental effects in downstream systems. In the case of estimating population-level health trends, for instance, we have already seen machine learning classifiers produce outcomes that are inconsistent across computational studies \cite{wolohan2020estimating,biester2021understanding,harrigian2022shift} and in conflict with traditional measurement techniques \cite{amir2019mental}. Continuing to pursue this line of research without questioning the validity of its underlying data has the potential to irreparably damage the public's trust in this domain, and worse, enable ill-informed decision making in highly-sensitive circumstances.


\section*{Acknowledgements}

We thank Ayah Zirikly and Carlos Aguirre for contributing annotations to use for evaluating inter-rater reliability. We also thank the anonymous reviewers for providing additional clinical grounding of our study and highlighting opportunities to improve our technical approach.


\bibliography{custom}


\appendix

\section{Interpreting Model Performance} \label{apx:interpreting}

\subsection{Data} \label{apx:interpretingdata}

Three individuals (one author $A_{1}$, two non-authors $B_{1},B_{2}$) independently generated the annotations used to facilitate the analysis presented in \S \ref{sec:Qualitative}. Statistics presented in the analysis are computed using the author's annotations, while reliability measures are computed using additional annotations from the non-authors. All annotators have several years of experience modeling language within social media to assess mental health, but do not claim to be experts in clinical psychology. Additionally, all annotators have prior experience with the CLPsych 2015 Shared Task data \citep{coppersmith2015clpsych} --- e.g., $A_{1}$ and $B_{1}$ have worked with the original CLPsych dataset extensively over the prior three years. We include the distribution of instances reviewed by each of our annotators in Table \ref{tab:annotsampledist}.

\begin{table} [h]
    \centering
    \begin{tabular}{c c c c c}
        & \multicolumn{3}{c}{\textbf{Time Period}} & \\ \cmidrule{2-4}
         & 2012-2015 & 2015-2018 & 2018-2021 & \textbf{Total}  \\ \toprule
        $A_{1}$ & 298 & 157 & 119 & 574 \\
        $B_{1}$ & 103 & 62 & 40 & 205 \\
        $B_{2}$ & 26 & 15 & 12 & 53 \\
    \end{tabular}
    \caption{Distribution of instances coded by each annotator across the three time periods. Note that the set of instances annotated follows the relationship: $B_2 \subseteq B_1 \subseteq A_1$.}
    \label{tab:annotsampledist}
\end{table}

\subsection{Inter-rater Reliability} \label{apx:intepretreliability}

\begin{figure}[t]
    \centering
    \includegraphics[width=\columnwidth]{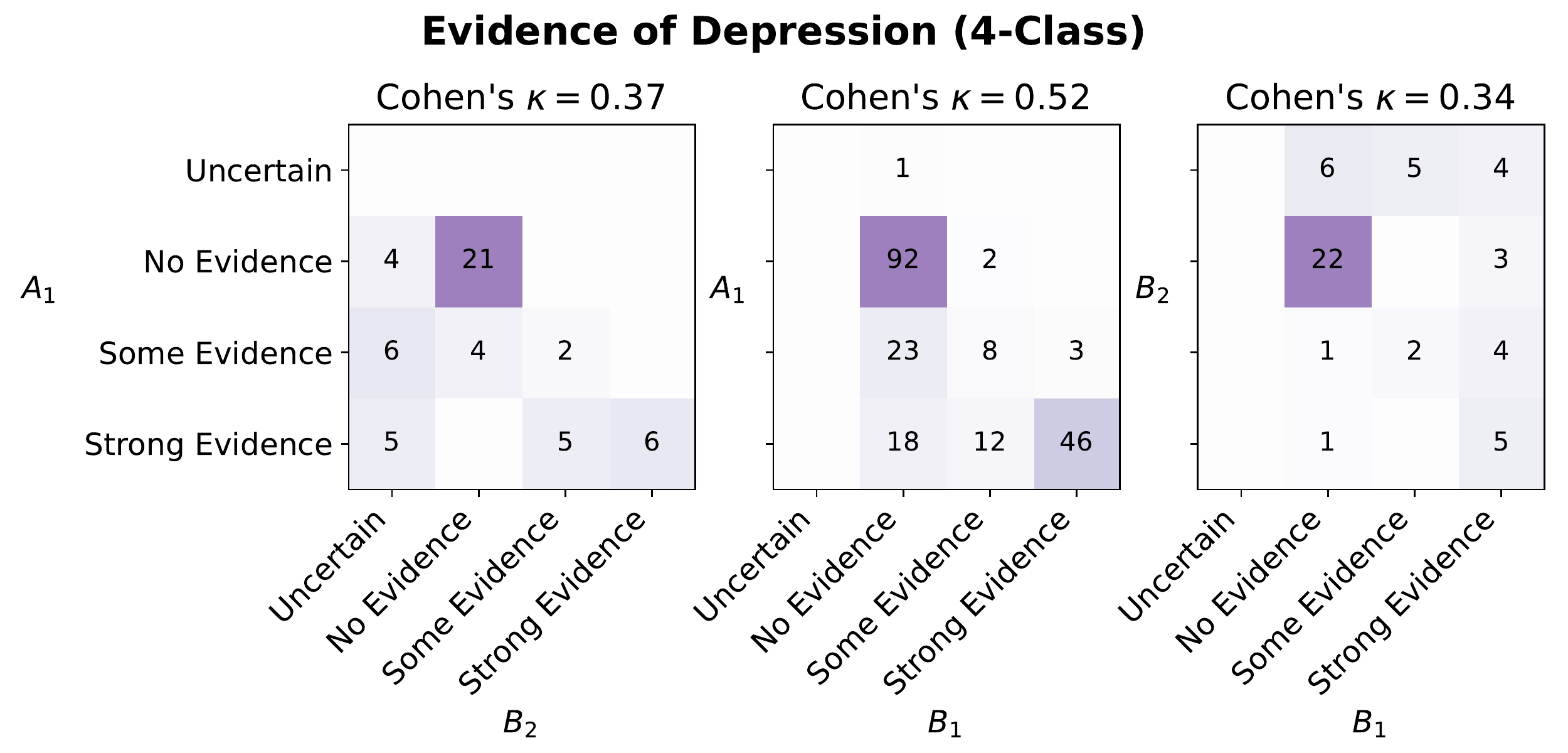}
    \quad
    \includegraphics[width=\columnwidth]{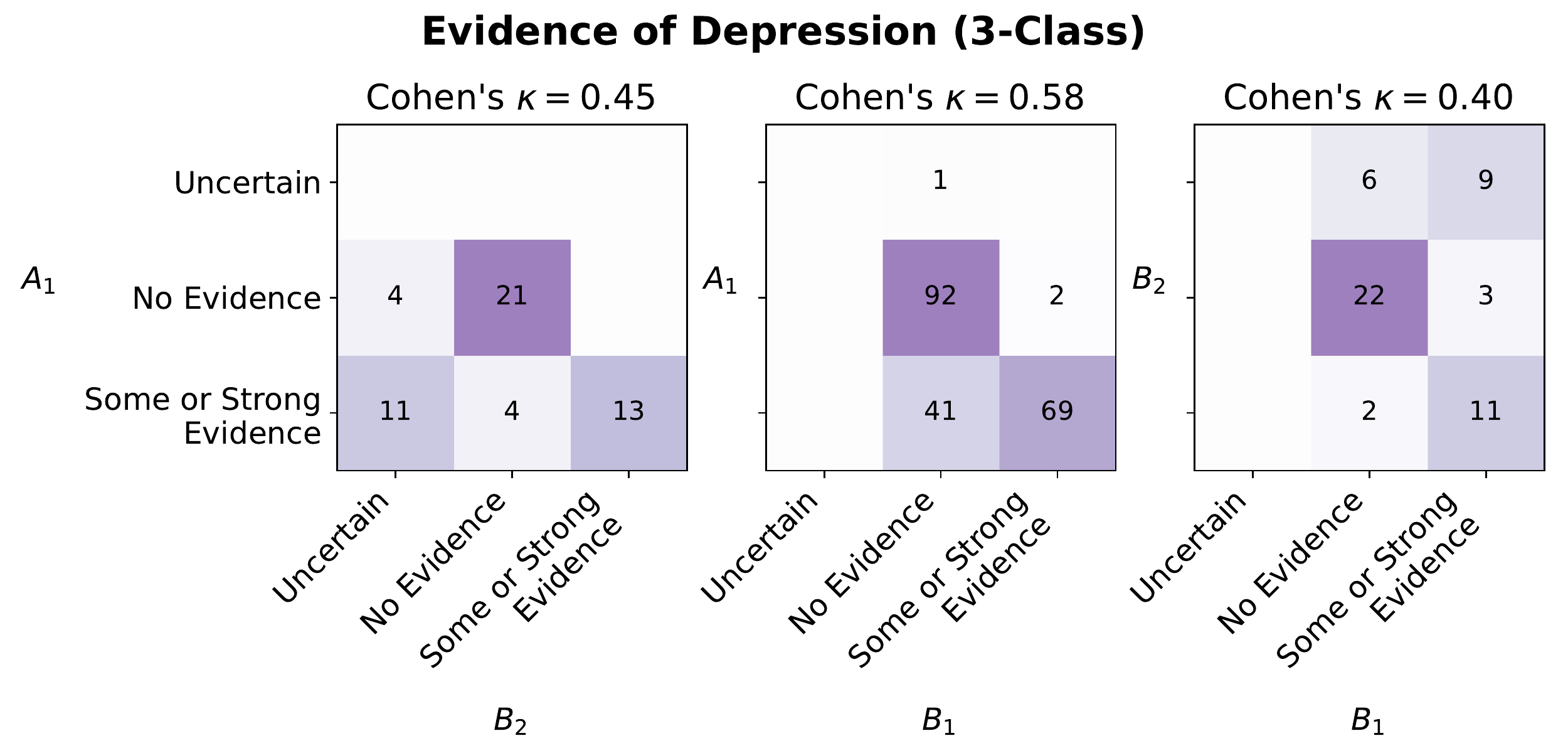}
    \quad
    \includegraphics[width=\columnwidth]{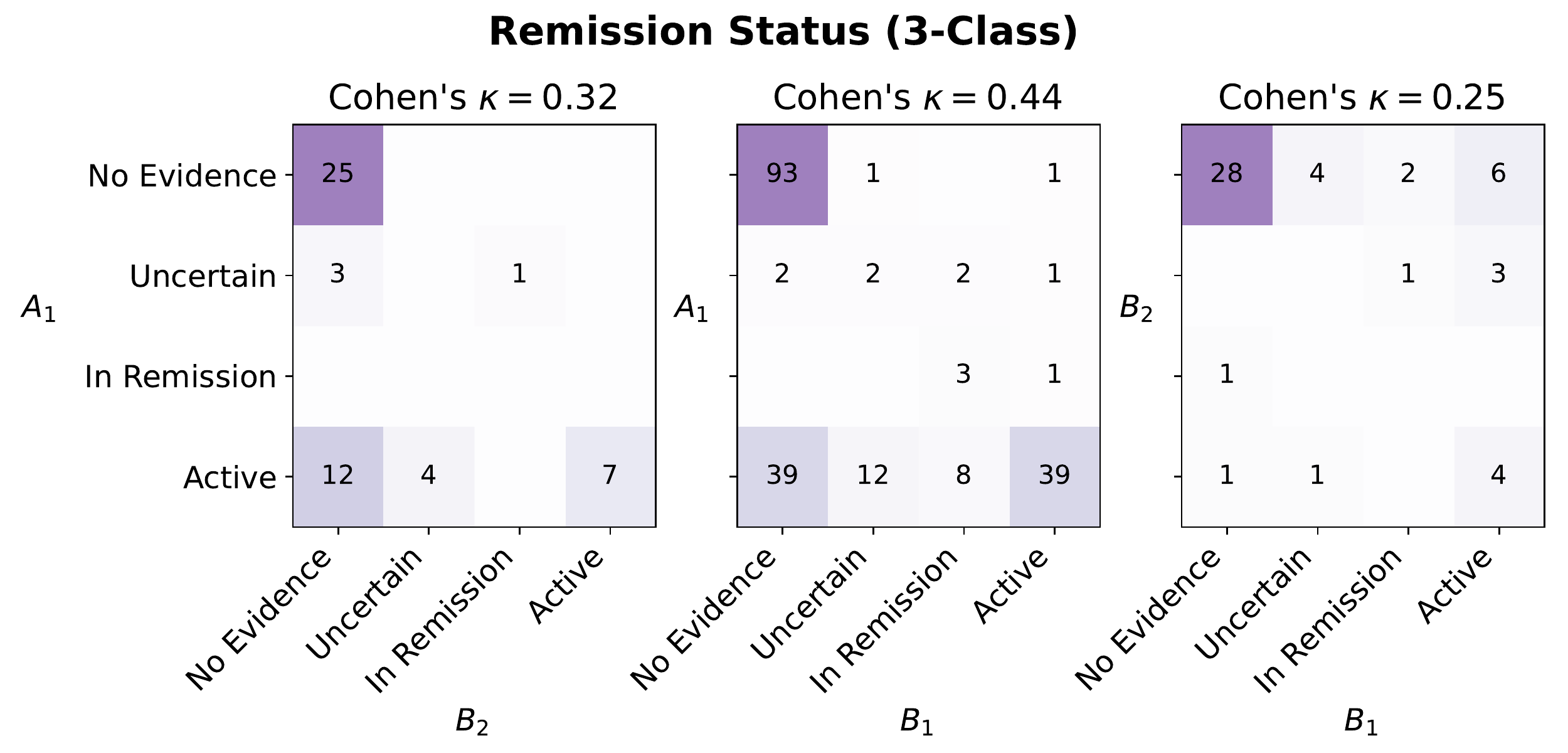}
    \caption{Pairwise agreement matrices for the annotation tasks. Underlying relationships reveal cognitive biases from annotator $A_{1}$ that may affect the outcomes presented in \S \ref{sec:Qualitative}.}
    \label{fig:pairwiseag}
\end{figure}

As a first look into inter-rater reliability, we consider three dimensions of agreement --- evidence of depression (four-class and three-class)\footnote{Note that the three-class evidence-of-depression grouping simply merges the Some Evidence and Strong Evidence categories of the four-class version.} and remission status (four-class). We present pairwise annotator agreement matrices for each of these dimensions in Figure \ref{fig:pairwiseag}. We use Cohen's kappa $\kappa$ to evaluate pairwise annotator agreement \citep{cohen1960coefficient} and Krippendorff's alpha $\alpha$ to evaluate multi-annotator agreement \citep{krippendorff2011computing}. 

We observe fair to moderate agreement for the evidence-of-depression task: $\alpha=0.4376$ and $\alpha=0.4988$ for the four-class and three-class versions, respectively. Meanwhile, agreement on remission status is poor, reflected by a Krippendorff's $\alpha$ of $0.3561$. In isolation, these agreement measures would suggest the results of our analysis should be accepted tentatively at best \citep{krippendorff2004reliability}. However, we argue these statistics are perhaps a bit conservative and skewed by the small sample size of annotations generated by $B_{2}$. A review of the underlying distributions provides us an opportunity to understand axes of disagreement and, in turn, contextualize the results presented in \S \ref{sec:Qualitative}.

As shown in Figure \ref{fig:pairwiseag}, annotator $B_{2}$ exhibits a higher propensity to use the ``Uncertain'' label in the evidence-of-depression tasks compared to annotators $A_{1}$ and $B_{1}$. At the same time, while annotator $B_{2}$ is more inclined to indicate they are uncertain about an example than annotator $A_{1}$, we note that annotator $B_{1}$ appears to have a higher baseline threshold of what constitutes evidence of depression than annotator $A_{1}$. The latter is demonstrated by the fact that nearly all examples marked in the affirmative by $B_{1}$ were also marked as such by $A_{1}$, but a large number of examples marked in the affirmative by $A_{1}$ were marked as not containing evidence of depression by $B_{1}$. 

With respect to the remission status task (bottom subplot of Figure \ref{fig:pairwiseag}), we note that annotator $B_{1}$ is more likely to mark an example as uncertain and more likely to mark an example as being in-remission than annotators $A_{1}$ and $B_{2}$. Broadly, this distribution highlights the difficulty of distinguishing active cases of clinical depression from prior experiences and lingering effects. It also serves as support for our recommendation in \S \ref{sec:Discussion} that researchers should attempt to include the time a diagnosis was received by an individual when curating new datasets. 

\begin{figure}[t]
    \centering
    \includegraphics[width=\columnwidth]{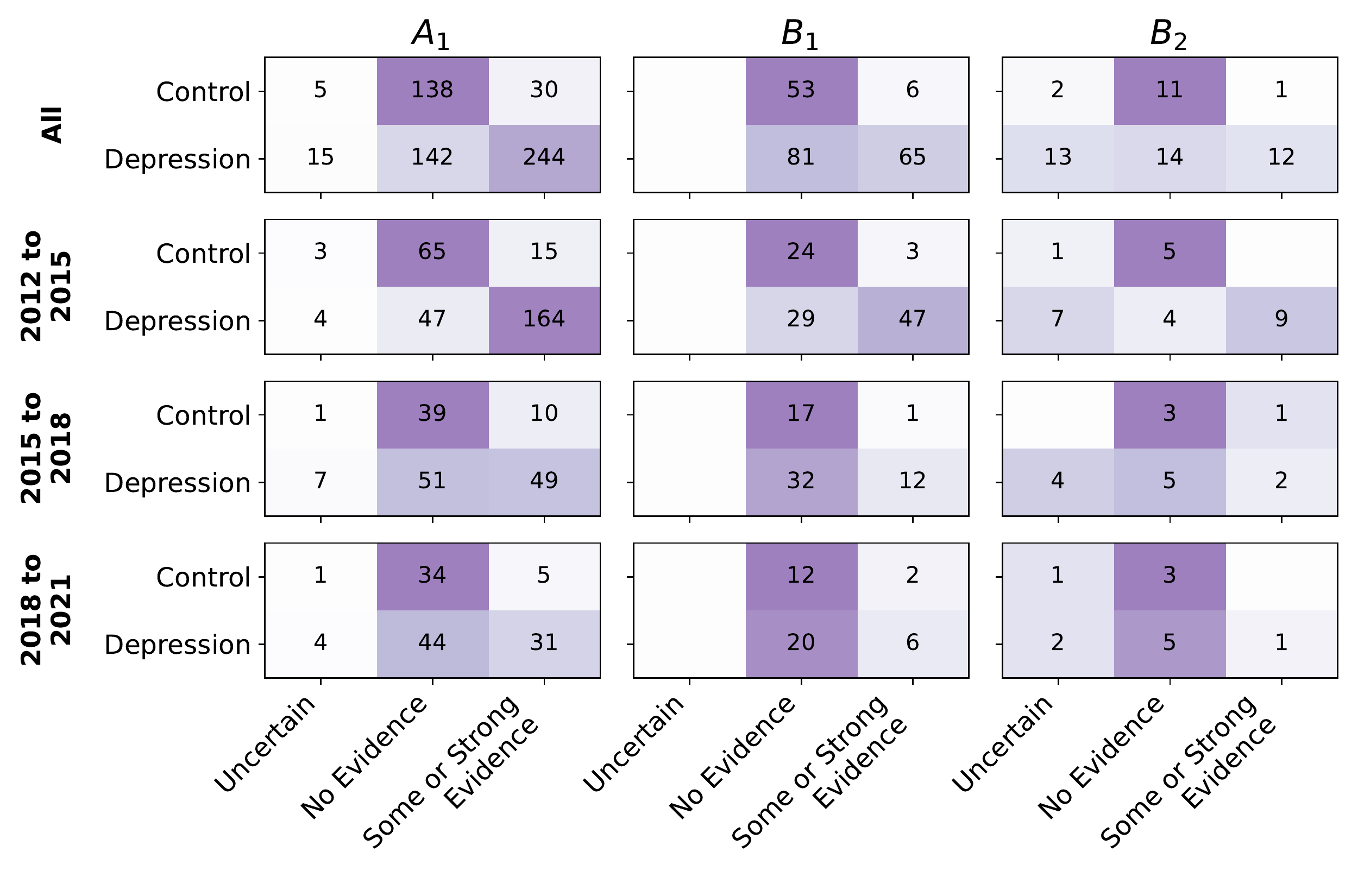}
    \caption{Distribution of annotations for the evidence of depression task (three-class) as a function of the original CLPsych labels. Affirmative evidence becomes less prevalent in the new time periods compared to the original time period for each annotator.}
    \label{fig:annotatelabel}
\end{figure}

We acquire additional context for our results by examining the distribution of annotations as a function of the original CLPsych labels. Examining the results visualized in Figure \ref{fig:annotatelabel}, we first note that annotator $A_{1}$ classifies instances most accurately (under the assumption that ground truth is fixed over time). We believe this outcome to be a result of exposure bias; the annotation task was conducted \emph{after} the completion of several modeling experiments, through which annotator $A_{1}$ was uniquely provided an opportunity to learn more about the presentation of depression by individuals in the 2015 CLPsych Shared Task dataset. We also note the distribution of ``Uncertain'' decisions from annotator $B_{2}$ concentrating within the original depression group. This seems to suggest annotator $B_{2}$ adopted a conservative coding approach when presented with instances that contained smaller degrees of evidence, whereas annotators $A_{1}$ and $B_{1}$ required a lower threshold of evidence to make a decision.

To conclude our reliability analysis, we examine agreement regarding the manner in which each annotator made their decision (i.e., evidence identification). We find that annotators $A_{1}$ and $B_{1}$ generally identify diagnosis disclosures within the same instances. Annotator $B_{2}$ often abstained from making a decision when presented with a disclosure due to uncertainty regarding the subject of the diagnosis. Annotator $A_{1}$ also indicated the presence of a depressed and/or irritable mood at a significantly higher rate than the other annotators, seemingly more sensitive to extreme negative emotions than the other annotators.

\textbf{Discussion.} Considering the difficulty of the annotation task, it is perhaps not surprising to have observed less than perfect annotator agreement. Machine learning classifiers often require hundreds of posts to make an accurate estimate of an individual's mental health status, while our annotators were only provided at maximum of 30 posts and encouraged to rely on varying levels of prior knowledge regarding the presentation of depression in social media. Critically, we emphasize that the goal of the analysis presented in \S \ref{sec:Qualitative} is \emph{not} to curate ground truth labels of mental health status or act as clinical experts, but rather to understand biases that may exist in a depression dataset generated using self-disclosed diagnoses. The analysis of inter-rater reliability presented above provides an opportunity to further ground the results discussed in \S \ref{sec:Qualitative} and highlight areas that may benefit from future research.

\subsection{Evidence Distribution} \label{apx:interpretingevidence}

\begin{figure*}[h!]
    \centering
    \includegraphics[width=\linewidth]{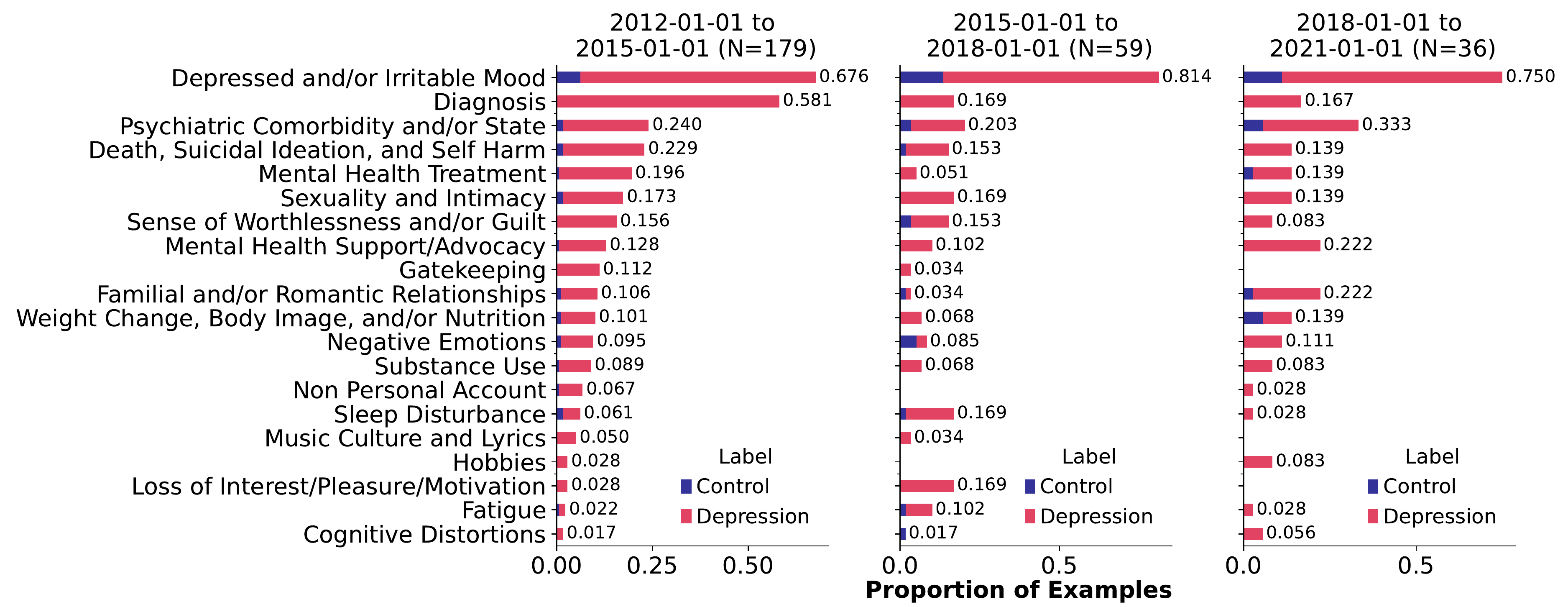}
    \caption{Distribution of evidence amongst individuals indicated as displaying at least some evidence of a depression diagnosis. Depressed and/or irritable mood is consistently the most common type of evidence within each of the three time periods.}
    \label{fig:Evidence}
\end{figure*}

We include a breakdown of evidence annotations for individuals displaying some evidence of depression (\S \ref{sec:Qualitative}) in Figure \ref{fig:Evidence}. Exemplary tweets for each of the evidence categories (paraphrased to maintain anonymity) are provided in Table \ref{tab:Examples}. Both can be found on the following pages.

\begin{table*}[!ht]
    \centering
    \def\arraystretch{0.925}%
    \begin{tabular}{l l} \toprule
    \textbf{Evidence} & 
        \textbf{Exemplary Tweets} \\ \toprule
        Diagnosis Disclosure & 
            ``Bipolar disorder and depression. My doctor finally agrees.'' \\&
            ``I have suffered from depression for several years now''
        \\ \midrule
        Depressed \& Irritable Mood & 
            ``No one ever asks if I'm doing fine.'' \\&
            ``You don't understand what I'm dealing with. Get fucked.''
        \\
        Loss of Interest/Pleasure/Motivation & 
            \begin{tabular}[t]{@{}l@{}}``...realizing you don't care about the things you used to enjoy''\end{tabular} \\&
            ``cant get out of bed today''
        \\
        Weight, Body Image, \& Nutrition &
            \begin{tabular}[t]{@{}l@{}}``Not that anyone cares, but I'm almost at my goal weight.''\end{tabular}\\&
            \begin{tabular}[t]{@{}l@{}}``I bought the dress I've always wanted, but still don't feel pretty.''\end{tabular}
        \\ 
        Sleep Disturbance & 
            \begin{tabular}[t]{@{}l@{}}``I CANT SLEEP. PAIN. JUST LIKE ALWAYS.''\end{tabular}\\&
            \begin{tabular}[t]{@{}l@{}}``Shit! Surviving on only a couple of hours of sleep again :/''\end{tabular}
        \\
        Fatigue & 
            \begin{tabular}[t]{@{}l@{}}``mentally drained from this pandemic''\end{tabular}\\&
            \begin{tabular}[t]{@{}l@{}}``This should be effortless but I can't work any harder''\end{tabular}
        \\
        Sense of Worthlessness \& Guilt & 
            \begin{tabular}[t]{@{}l@{}}``when you let someone do anything to you...''\end{tabular}\\&
            \begin{tabular}[t]{@{}l@{}}``It truly is always my fault. I probably suck.''\end{tabular}
        \\
        Impaired Thought &
            \begin{tabular}[t]{@{}l@{}}``I'm failing my classes because I'm depressed.''\end{tabular}\\&
            \begin{tabular}[t]{@{}l@{}}``at work. cant focus doe''\end{tabular}
        \\
        Death \& Self Harm & 
            \begin{tabular}[t]{@{}l@{}}``My scars are faded...unless you care to look close''\end{tabular}\\&
            \begin{tabular}[t]{@{}l@{}}``I wish you all never see a loved one fade away.''\end{tabular}
        \\ \midrule
        Cognitive Distortions &
            \begin{tabular}[t]{@{}l@{}}``Going to fail this exam. SCREWED.''\end{tabular}\\&
            \begin{tabular}[t]{@{}l@{}}``I always think my bf is going to leave me''\end{tabular}
        \\
        Treatment & 
            \begin{tabular}[t]{@{}l@{}}``Scared to tell a women that I'm in therapy''\end{tabular}\\&
            \begin{tabular}[t]{@{}l@{}}``Slowly weaning of the prozac.''\end{tabular}
        \\
        Gatekeeping & 
            \begin{tabular}[t]{@{}l@{}}``depression isn't just a bad day. fuck you all.''\end{tabular}\\&
            \begin{tabular}[t]{@{}l@{}}``LET ME SHOW YOU WANT DEPRESSION IS''\end{tabular}
        \\
        Sexuality and Intimacy & 
            \begin{tabular}[t]{@{}l@{}}``Who wants to come take some pics of me for only fans? ;)''\end{tabular}\\&
            \begin{tabular}[t]{@{}l@{}}``Every girl should watch porn with their bf''\end{tabular}
        \\
        Negative Emotions & 
            \begin{tabular}[t]{@{}l@{}}``hi sunshine! Too bad no one to spend today with.''\end{tabular}\\&
            \begin{tabular}[t]{@{}l@{}}``I feel like no one cares even though I know they do''\end{tabular}
        \\
        Coping Strategies & 
            \begin{tabular}[t]{@{}l@{}}``Have you talked to anyone about it yet?''\end{tabular}\\&
            \begin{tabular}[t]{@{}l@{}}``Art is always the easiest way to distract me from my anxiety''\end{tabular}
        \\
        Psychiatric Comorbidity \& State &
            \begin{tabular}[t]{@{}l@{}}``Really stressing today. Lots of built up anger''\end{tabular}\\&
            \begin{tabular}[t]{@{}l@{}}``I am anorexic and cut myself.''\end{tabular}
        \\
        Non-psychiatric Comorbidity & 
            \begin{tabular}[t]{@{}l@{}}``Could use a little bit of aid \#DisabilityAid''\end{tabular}\\&
            \begin{tabular}[t]{@{}l@{}}``Lots of back pain ruining what should be a beautiful day.''\end{tabular}
        \\
        Substance Use & 
            \begin{tabular}[t]{@{}l@{}}``I really shouldn't be drunk this early.''\end{tabular}\\&
            \begin{tabular}[t]{@{}l@{}}``Weed makes the dreams go away and thats a good thing.''\end{tabular}
        \\
        Support \& Advocacy & 
            \begin{tabular}[t]{@{}l@{}}``If I can manage a smile, I believe you can too one day!''\end{tabular}\\&
            \begin{tabular}[t]{@{}l@{}}``RIP Chester. If you're going through pain, reach out to me.''\end{tabular}
        \\
        Personality and Identity & 
            \begin{tabular}[t]{@{}l@{}}``Girls say they love a man in uniform until they do their job''\end{tabular}\\&
            \begin{tabular}[t]{@{}l@{}}``Lol grandma still think I'm bringing a boy home''\end{tabular}
        \\
        Music Culture \& Lyrics &
            \begin{tabular}[t]{@{}l@{}}``\#FallingInReverse :D''\end{tabular}\\&
            \begin{tabular}[t]{@{}l@{}}``Scene doesn't mean emo idiots. I dont want to kill myself.''\end{tabular}
        \\
        Familial/Romantic Relationships & 
                    \begin{tabular}[t]{@{}l@{}}``when bae dont answer the phone xx''\end{tabular}\\&
                    \begin{tabular}[t]{@{}l@{}}``Mom: You'll never lose weight. Me: Is that why dad left?''\end{tabular}
        \\
        Political \& Moral Beliefs & 
            \begin{tabular}[t]{@{}l@{}}``look in the mirror if you're not upset a cop can murder''\end{tabular}\\&
            \begin{tabular}[t]{@{}l@{}}``Trump will kill us all''\end{tabular}
        \\
        Hobbies & 
            \begin{tabular}[t]{@{}l@{}}``Missin the old days when eveyone played Pokemon yellow''\end{tabular}\\&
            \begin{tabular}[t]{@{}l@{}}``Boys that watch the Kardashians. Love.''\end{tabular}
        \\
        Non-personal Accounts &
             \begin{tabular}[t]{@{}l@{}}``My life was about to fall apart until I found the Calm app...''\end{tabular}\\&
             \begin{tabular}[t]{@{}l@{}}``Breaking News: 5-alarm fire just outside Tulsa...''\end{tabular} \\
    \bottomrule
    \end{tabular}
    \caption{Exemplary tweets and phrases (modified to preserve anonymity) for each of the 25 evidence categories.}
    \label{tab:Examples}
\end{table*}

\end{document}